\documentclass[sigconf]{acmart}

\pdfoutput=1

\usepackage{multirow}

\usepackage{siunitx}
\usepackage{dsfont}
\usepackage{pifont}% http://ctan.org/pkg/pifont
\usepackage{subcaption}
\usepackage{pgfplots}
\usepackage{bm}
\newcommand{\cmark}{{\color{green}\ding{51}}}%
\newcommand{\xmark}{{\color{red}\ding{55}}}%

\usepackage{tikzit}
\tikzstyle{circle}=[fill=white, draw=black, shape=circle]
\tikzstyle{color1}=[fill=none, draw=none, shape=circle, text={rgb,255: red,3; green,152; blue,168}, align=left]
\tikzstyle{color2}=[fill=none, draw=none, shape=circle, text={rgb,255: red,255; green,171; blue,64}, align=left]
\tikzstyle{color3}=[fill=none, draw=none, shape=circle, text={rgb,255: red,204; green,65; blue,37}, align=left]
\tikzstyle{left}=[fill=none, draw=none, shape=rectangle, align=left]
\tikzstyle{centerGrey}=[fill={rgb,255: red,238; green,238; blue,238}, draw=none, shape=rectangle, align=center]
\tikzstyle{rotateBold}=[fill=none, draw=none, shape=circle, rotate=90]
\tikzstyle{rotate45}=[fill=none, draw=none, shape=circle, rotate=45, align=left]
\tikzstyle{circleB}=[fill=white, draw=black, shape=circle, line width=1.5pt]
\tikzstyle{circleB1}=[fill=white, draw={rgb,255: red,3; green,152; blue,168}, shape=circle, line width=1.5pt]
\tikzstyle{circleB2}=[fill=white, draw={rgb,255: red,255; green,171; blue,64}, shape=circle, line width=1.5pt]
\tikzstyle{circleB3}=[fill=white, draw={rgb,255: red,204; green,65; blue,37}, shape=circle, line width=1.5pt]
\tikzstyle{rotateLeft}=[fill=none, draw=none, shape=rectangle, rotate=90, align=center]

\tikzstyle{grey edge}=[-, fill={rgb,255: red,238; green,238; blue,238}, draw={rgb,255: red,235; green,235; blue,235}]
\tikzstyle{black arrow}=[->, line width=2pt]
\tikzstyle{dotted1}=[-, dashed, draw={rgb,255: red,3; green,152; blue,168}, line width=1pt]
\tikzstyle{dotted2}=[-, dashed, draw={rgb,255: red,255; green,171; blue,64}, line width=1pt]
\tikzstyle{dotted3}=[-, dashed, draw={rgb,255: red,204; green,65; blue,37}, line width=1pt]
\tikzstyle{dotted grey}=[-, dashed, draw={rgb,255: red,180; green,180; blue,180}, line width=0.5pt]
\tikzstyle{grey edge only}=[-, draw={rgb,255: red,180; green,180; blue,180}]
\tikzstyle{blackArrow}=[->, line width=0.5pt]
\tikzstyle{solid1}=[-, draw={rgb,255: red,3; green,152; blue,168}, line width=1.5pt]
\tikzstyle{solid2}=[-, draw={rgb,255: red,255; green,171; blue,64}, line width=1.5pt]
\tikzstyle{solid3}=[-, draw={rgb,255: red,204; green,65; blue,37}, line width=1.5pt]
\tikzstyle{solid1Fill}=[-, draw={rgb,255: red,3; green,152; blue,168}, line width=1.5pt, fill=white]
\tikzstyle{solid2Fill}=[-, draw={rgb,255: red,255; green,171; blue,64}, fill=white, line width=1.5pt]
\tikzstyle{solid3Fill}=[-, draw={rgb,255: red,204; green,65; blue,37}, line width=1.5pt, fill=white]
\tikzstyle{arrowDottedGray}=[->, line width=0.5pt, draw={rgb,255: red,180; green,180; blue,180}, dashed]
\tikzstyle{solidB}=[-, line width=1.5pt, draw=black, fill=white]
\tikzstyle{black}=[-, line width=1.5pt, draw=black]

\AtBeginDocument{%
  }

\copyrightyear{2023}
\acmYear{2023}
\setcopyright{rightsretained}
\acmConference[ICMI '23]{INTERNATIONAL CONFERENCE ON MULTIMODAL INTERACTION}{October 9--13, 2023}{Paris, France}
\acmBooktitle{INTERNATIONAL CONFERENCE ON MULTIMODAL INTERACTION (ICMI '23), October 9--13, 2023, Paris, France}\acmDOI{10.1145/3577190.3614115}
\acmISBN{979-8-4007-0055-2/23/10}

\begin{document}

\sisetup{detect-all=true}

\title{Neural Mixed Effects for Nonlinear Personalized Predictions}

\author{Torsten Wörtwein}
\email{twoertwe@cs.cmu.edu}
\orcid{0009-0003-5659-029X}
\affiliation{%
  \institution{Carnegie Mellon University}
  \streetaddress{5000 Forbes Ave}
  \city{Pittsburgh}
  \state{PA}
  \country{USA}
  \postcode{15213}
}

\author{Nicholas B. Allen}
\email{nallen3@uoregon.edu}
\orcid{0000-0002-1086-6639}
\affiliation{%
  \institution{University of Oregon}
  \streetaddress{1585 E 13th Ave}
  \city{Eugene}
  \state{OR}
  \country{USA}
  \postcode{97403}}

\author{Lisa B. Sheeber}
\email{lsheeber@ori.org}
\orcid{0000-0003-1293-943X}
\affiliation{%
  \institution{Oregon Research Institute}
  \streetaddress{3800 Sports Way}
  \city{Springfield}
  \state{OR}
  \country{USA}
  \postcode{97477}}

\author{Randy P. Auerbach}
\email{rpa2009@cumc.columbia.edu}
\orcid{0000-0003-2319-4744}
\affiliation{%
  \institution{Columbia University}
  \streetaddress{2960 Broadway}
  \city{New York}
  \state{NY}
  \country{USA}
  \postcode{10027}}

\author{Jeffrey F. Cohn}
\email{jeffcohn@pitt.edu}
\orcid{0000-0002-9393-1116}
\affiliation{%
  \institution{University of Pittsburgh}
  \streetaddress{4200 Fifth Ave}
  \city{Pittsburgh}
  \state{PA}
  \country{USA}
  \postcode{15260}}

\author{Louis-Philippe Morency}
\email{morency@cs.cmu.edu}
\orcid{0000-0001-6376-7696}
\affiliation{%
  \institution{Carnegie Mellon University}
  \streetaddress{ 5000 Forbes Ave}
  \city{Pittsburgh}
  \state{PA}
  \country{USA}
  \postcode{15213}}

\renewcommand{\shortauthors}{Wörtwein et al.}

\begin{abstract}
Personalized prediction is a machine learning approach that predicts a person's future observations based on their past labeled observations and is typically used for sequential tasks, e.g., to predict daily mood ratings. When making personalized predictions, a model can combine two types of trends: (a) trends shared across people, i.e., person-generic trends, such as being happier on weekends, and (b) unique trends for each person, i.e., person-specific trends, such as a stressful weekly meeting. Mixed effect models are popular statistical models to study both trends by combining person-generic and person-specific parameters. Though linear mixed effect models are gaining popularity in machine learning by integrating them with neural networks, these integrations are currently limited to linear person-specific parameters: ruling out nonlinear person-specific trends. In this paper, we propose Neural Mixed Effect (NME) models to optimize nonlinear person-specific parameters anywhere in a neural network in a scalable manner\footnote{Our code is publicly available at \url{https://github.com/twoertwein/NeuralMixedEffects}.}. NME combines the efficiency of neural network optimization with nonlinear mixed effects modeling. Empirically, we observe that NME improves performance across six unimodal and multimodal datasets, including a smartphone dataset to predict daily mood and a mother-adolescent dataset to predict affective state sequences where half the mothers experience symptoms of depression. Furthermore, we evaluate NME for two model architectures, including for neural conditional random fields (CRF) to predict affective state sequences where the CRF learns nonlinear person-specific temporal transitions between affective states. Analysis of these person-specific transitions on the mother-adolescent dataset shows interpretable trends related to the mother's depression symptoms.
\end{abstract}




\begin{teaserfigure}
  \centering
  \resizebox{\textwidth}{!}{\input{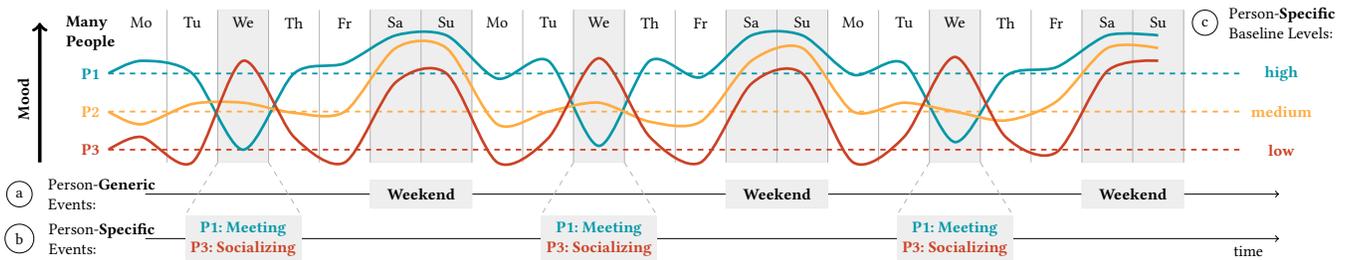}}
  \caption{Illustration of why combining both person-generic and person-specific trends is important when learning personalized prediction models. The illustrated example is for daily mood prediction. (a) Most people are happier on weekends when they do not have to work. (b) Specific individuals, in our case P1 and P3, may have weekly events impacting their mood, e.g., socializing with friends can be positive, while a stressful meeting can be negative. (c) It is important to further know the baseline mood level of each person, as it varies between people, as shown for P1, P2, and P3.}
  \Description{Contour of daily mood levels over three weeks with markers of weekends and person-specific events.}
  \label{fig:problem}
\end{teaserfigure}


\maketitle

\section{Introduction}

\begin{figure*}
    \centering
    \begin{subfigure}[t]{.24\textwidth}
            \resizebox{.9\textwidth}{!}{\begin{tikzpicture}
    \clip(-3.4,0.25) rectangle (0.8,7.9);
	\begin{pgfonlayer}{nodelayer}
		\node [style=none] (74) at (-2.9, 6.15) {};
		\node [style=none] (75) at (-2.9, 5.025) {};
		\node [style=none] (76) at (0.4, 6.15) {};
		\node [style=none] (77) at (0.4, 5.025) {};
		\node [style=circleB] (78) at (-2.75, 5.275) {};
		\node [style=circleB3] (79) at (-2.625, 5.875) {};
		\node [style=circleB2] (80) at (-2.75, 5.875) {};
		\node [style=circleB1] (81) at (-2.875, 5.875) {};
		\node [style=none] (82) at (-2.75, 5.575) {\small +};
		\node [style=circleB] (83) at (-1.75, 5.275) {};
		\node [style=circleB3] (84) at (-1.625, 5.875) {};
		\node [style=circleB2] (85) at (-1.75, 5.875) {};
		\node [style=circleB1] (86) at (-1.875, 5.875) {};
		\node [style=none] (87) at (-1.75, 5.575) {\small +};
		\node [style=circleB] (88) at (-0.75, 5.275) {};
		\node [style=circleB3] (89) at (-0.625, 5.875) {};
		\node [style=circleB2] (90) at (-0.75, 5.875) {};
		\node [style=circleB1] (91) at (-0.875, 5.875) {};
		\node [style=none] (92) at (-0.75, 5.575) {\small+};
		\node [style=circleB] (93) at (0.25, 5.275) {};
		\node [style=circleB3] (94) at (0.375, 5.875) {};
		\node [style=circleB2] (95) at (0.25, 5.875) {};
		\node [style=circleB1] (96) at (0.125, 5.875) {};
		\node [style=none] (97) at (0.25, 5.575) {\small +};
		\node [style=none] (98) at (-1.25, 6.15) {};
		\node [style=none] (99) at (-1.25, 5.025) {};
		\node [style=none] (108) at (-2.75, 4.9) {};
		\node [style=none] (109) at (-2, 4.9) {};
		\node [style=none] (110) at (-0.5, 4.9) {};
		\node [style=none] (111) at (0.25, 4.9) {};
		\node [style=none] (112) at (-2.75, 4.4) {};
		\node [style=none] (113) at (-2, 4.4) {};
		\node [style=none] (114) at (-0.5, 4.4) {};
		\node [style=none] (115) at (0.25, 4.4) {};
		\node [style=none] (116) at (-1.25, 4.9) {};
		\node [style=none] (117) at (-1.25, 4.4) {};
		\node [style=none] (118) at (-1.25, 4.15) {Input Features};
		\node [style=none] (120) at (-1.625, 7.375) {};
		\node [style=none] (124) at (-1.625, 6.875) {};
		\node [style=none] (127) at (-0.875, 7.375) {};
		\node [style=none] (128) at (-0.875, 6.875) {};
		\node [style=none] (129) at (-1.25, 7.625) {Output Label};
		\node [style=none] (130) at (-1.25, 6.875) {};
		\node [style=circleB] (131) at (0.275, 1.775) {};
		\node [style=circleB3] (132) at (0.4, 0.775) {};
		\node [style=circleB2] (133) at (0.275, 0.775) {};
		\node [style=circleB1] (134) at (0.15, 0.775) {};
		\node [style=left] (135) at (-1.6, 1.775) {Person-\textbf{Generic}\\Parameters ($\bar{\theta}$)};
		\node [style=left] (136) at (-1.6, 0.775) {Person-\textbf{Specific}\\Parameters ($\theta^i$)};
		\node [style=left] (137) at (-2.6, 2.775) {Legend};
	\end{pgfonlayer}
	\begin{pgfonlayer}{edgelayer}
		\draw (75.center)
			 to [bend left=90, looseness=1.25] (74.center)
			 to (76.center)
			 to [bend left=90, looseness=1.25] (77.center)
			 to cycle;
		\draw [style=grey edge only] (108.center) to (112.center);
		\draw [style=grey edge only] (112.center) to (115.center);
		\draw [style=grey edge only] (115.center) to (111.center);
		\draw [style=grey edge only] (111.center) to (108.center);
		\draw [style=grey edge only] (109.center) to (113.center);
		\draw [style=grey edge only] (110.center) to (114.center);
		\draw [style=grey edge only] (116.center) to (117.center);
		\draw [style=grey edge only] (120.center) to (124.center);
		\draw [style=grey edge only] (127.center) to (128.center);
		\draw [style=blackArrow] (116.center) to (99.center);
		\draw [style=blackArrow] (98.center) to (130.center);
		\draw [style=grey edge only] (128.center) to (124.center);
		\draw [style=grey edge only] (127.center) to (120.center);
	\end{pgfonlayer}
\end{tikzpicture}}
            \caption{Linear Mixed Effects (LME)~\cite{lindstrom1988newton}}
            \label{fig:lme}
        \end{subfigure}\hfill%   
        \begin{subfigure}[t]{.24\textwidth}
            \resizebox{.9\textwidth}{!}{\begin{tikzpicture}
    \clip(-3.4,0.25) rectangle (0.8,7.9);
	\begin{pgfonlayer}{nodelayer}
		\node [style=none] (74) at (-2.9, 6.15) {};
		\node [style=none] (75) at (-2.9, 5.025) {};
		\node [style=none] (76) at (0.4, 6.15) {};
		\node [style=none] (77) at (0.4, 5.025) {};
		\node [style=circleB] (78) at (-2.75, 5.275) {};
		\node [style=circleB3] (79) at (-2.625, 5.875) {};
		\node [style=circleB2] (80) at (-2.75, 5.875) {};
		\node [style=circleB1] (81) at (-2.875, 5.875) {};
		\node [style=none] (82) at (-2.75, 5.575) {\small +};
		\node [style=circleB] (83) at (-1.75, 5.275) {};
		\node [style=circleB3] (84) at (-1.625, 5.875) {};
		\node [style=circleB2] (85) at (-1.75, 5.875) {};
		\node [style=circleB1] (86) at (-1.875, 5.875) {};
		\node [style=none] (87) at (-1.75, 5.575) {\small +};
		\node [style=circleB] (88) at (-0.75, 5.275) {};
		\node [style=circleB3] (89) at (-0.625, 5.875) {};
		\node [style=circleB2] (90) at (-0.75, 5.875) {};
		\node [style=circleB1] (91) at (-0.875, 5.875) {};
		\node [style=none] (92) at (-0.75, 5.575) {\small+};
		\node [style=circleB] (93) at (0.25, 5.275) {};
		\node [style=circleB3] (94) at (0.375, 5.875) {};
		\node [style=circleB2] (95) at (0.25, 5.875) {};
		\node [style=circleB1] (96) at (0.125, 5.875) {};
		\node [style=none] (97) at (0.25, 5.575) {\small +};
		\node [style=none] (98) at (-1.25, 6.15) {};
		\node [style=none] (99) at (-1.25, 5.025) {};
		\node [style=none] (100) at (-2.85, 6.75) {};
		\node [style=none] (101) at (-2.85, 6.275) {};
		\node [style=none] (102) at (0.45, 6.75) {};
		\node [style=none] (103) at (0.45, 6.275) {};
		\node [style=none] (104) at (-1.25, 6.275) {};
		\node [style=none] (105) at (-1.25, 6.275) {};
		\node [style=none] (106) at (-1.25, 6.75) {};
		\node [style=none] (107) at (-1.25, 6.5) {Nonlinear Layer};
		\node [style=none] (108) at (-2.75, 4.9) {};
		\node [style=none] (109) at (-2, 4.9) {};
		\node [style=none] (110) at (-0.5, 4.9) {};
		\node [style=none] (111) at (0.25, 4.9) {};
		\node [style=none] (112) at (-2.75, 4.4) {};
		\node [style=none] (113) at (-2, 4.4) {};
		\node [style=none] (114) at (-0.5, 4.4) {};
		\node [style=none] (115) at (0.25, 4.4) {};
		\node [style=none] (116) at (-1.25, 4.9) {};
		\node [style=none] (117) at (-1.25, 4.4) {};
		\node [style=none] (118) at (-1.25, 4.15) {Input Features};
		\node [style=none] (119) at (-2.375, 7.375) {};
		\node [style=none] (120) at (-1.625, 7.375) {};
		\node [style=none] (121) at (-0.125, 7.375) {};
		\node [style=none] (123) at (-2.375, 6.875) {};
		\node [style=none] (124) at (-1.625, 6.875) {};
		\node [style=none] (125) at (-0.125, 6.875) {};
		\node [style=none] (127) at (-0.875, 7.375) {};
		\node [style=none] (128) at (-0.875, 6.875) {};
		\node [style=none] (129) at (-1.25, 7.625) {Output Labels};
		\node [style=none] (130) at (-1.25, 6.875) {};
	\end{pgfonlayer}
	\begin{pgfonlayer}{edgelayer}
		\draw (75.center)
			 to [bend left=90, looseness=1.25] (74.center)
			 to (76.center)
			 to [bend left=90, looseness=1.25] (77.center)
			 to cycle;
		\draw [bend left=90, looseness=1.25] (101.center) to (100.center);
		\draw (100.center) to (102.center);
		\draw [bend left=90, looseness=1.25] (102.center) to (103.center);
		\draw (103.center) to (101.center);
		\draw [style=blackArrow] (98.center) to (105.center);
		\draw [style=grey edge only] (108.center) to (112.center);
		\draw [style=grey edge only] (112.center) to (115.center);
		\draw [style=grey edge only] (115.center) to (111.center);
		\draw [style=grey edge only] (111.center) to (108.center);
		\draw [style=grey edge only] (109.center) to (113.center);
		\draw [style=grey edge only] (110.center) to (114.center);
		\draw [style=grey edge only] (116.center) to (117.center);
		\draw [style=grey edge only] (119.center) to (123.center);
		\draw [style=grey edge only] (120.center) to (124.center);
		\draw [style=grey edge only] (121.center) to (125.center);
		\draw [style=grey edge only] (127.center) to (128.center);
		\draw [style=grey edge only] (119.center) to (121.center);
		\draw [style=grey edge only] (123.center) to (125.center);
		\draw [style=blackArrow] (106.center) to (130.center);
		\draw [style=blackArrow] (116.center) to (99.center);
	\end{pgfonlayer}
\end{tikzpicture}}
            \caption{Nonlinear Mixed Effects (NLME)~\cite{saem}}
            \label{fig:nlme}
        \end{subfigure}\hfill%
        \begin{subfigure}[t]{.24\textwidth}
            \resizebox{.9\textwidth}{!}{\begin{tikzpicture}
    \clip(-3.4,0.25) rectangle (0.8,7.9);
	\begin{pgfonlayer}{nodelayer}
		\node [style=none] (0) at (-2.9, 3.075) {};
		\node [style=none] (1) at (-2.9, 2.575) {};
		\node [style=none] (2) at (0.4, 3.075) {};
		\node [style=none] (3) at (0.4, 2.575) {};
		\node [style=circleB] (4) at (-2.75, 2.825) {};
		\node [style=circleB] (15) at (-1.75, 2.825) {};
		\node [style=circleB] (20) at (-0.75, 2.825) {};
		\node [style=circleB] (25) at (0.25, 2.825) {};
		\node [style=none] (30) at (-1.25, 3.075) {};
		\node [style=none] (31) at (-1.25, 2.575) {};
		\node [style=none] (32) at (-2.85, 3.675) {};
		\node [style=none] (33) at (-2.85, 3.2) {};
		\node [style=none] (34) at (0.45, 3.675) {};
		\node [style=none] (35) at (0.45, 3.2) {};
		\node [style=none] (36) at (-1.25, 3.2) {};
		\node [style=none] (37) at (-1.25, 3.2) {};
		\node [style=none] (38) at (-1.25, 3.675) {};
		\node [style=none] (39) at (-1.25, 3.425) {Nonlinear Layer};
		\node [style=none] (40) at (-2.9, 4.3) {};
		\node [style=none] (41) at (-2.9, 3.8) {};
		\node [style=none] (42) at (0.4, 4.3) {};
		\node [style=none] (43) at (0.4, 3.8) {};
		\node [style=circleB] (44) at (-2.75, 4.05) {};
		\node [style=circleB] (49) at (-1.75, 4.05) {};
		\node [style=circleB] (54) at (-0.75, 4.05) {};
		\node [style=circleB] (59) at (0.25, 4.05) {};
		\node [style=none] (64) at (-1.25, 4.3) {};
		\node [style=none] (65) at (-1.25, 3.8) {};
		\node [style=none] (66) at (-2.85, 4.9) {};
		\node [style=none] (67) at (-2.85, 4.425) {};
		\node [style=none] (68) at (0.45, 4.9) {};
		\node [style=none] (69) at (0.45, 4.425) {};
		\node [style=none] (70) at (-1.25, 4.425) {};
		\node [style=none] (71) at (-1.25, 4.425) {};
		\node [style=none] (72) at (-1.25, 4.9) {};
		\node [style=none] (73) at (-1.25, 4.65) {Nonlinear Layer};
		\node [style=none] (74) at (-2.9, 6.15) {};
		\node [style=none] (75) at (-2.9, 5.025) {};
		\node [style=none] (76) at (0.4, 6.15) {};
		\node [style=none] (77) at (0.4, 5.025) {};
		\node [style=circleB] (78) at (-2.75, 5.275) {};
		\node [style=circleB3] (79) at (-2.625, 5.875) {};
		\node [style=circleB2] (80) at (-2.75, 5.875) {};
		\node [style=circleB1] (81) at (-2.875, 5.875) {};
		\node [style=none] (82) at (-2.75, 5.575) {\small +};
		\node [style=circleB] (83) at (-1.75, 5.275) {};
		\node [style=circleB3] (84) at (-1.625, 5.875) {};
		\node [style=circleB2] (85) at (-1.75, 5.875) {};
		\node [style=circleB1] (86) at (-1.875, 5.875) {};
		\node [style=none] (87) at (-1.75, 5.575) {\small +};
		\node [style=circleB] (88) at (-0.75, 5.275) {};
		\node [style=circleB3] (89) at (-0.625, 5.875) {};
		\node [style=circleB2] (90) at (-0.75, 5.875) {};
		\node [style=circleB1] (91) at (-0.875, 5.875) {};
		\node [style=none] (92) at (-0.75, 5.575) {\small+};
		\node [style=circleB] (93) at (0.25, 5.275) {};
		\node [style=circleB3] (94) at (0.375, 5.875) {};
		\node [style=circleB2] (95) at (0.25, 5.875) {};
		\node [style=circleB1] (96) at (0.125, 5.875) {};
		\node [style=none] (97) at (0.25, 5.575) {\small +};
		\node [style=none] (98) at (-1.25, 6.15) {};
		\node [style=none] (99) at (-1.25, 5.025) {};
		\node [style=none] (108) at (-2.75, 2.45) {};
		\node [style=none] (109) at (-2, 2.45) {};
		\node [style=none] (110) at (-0.5, 2.45) {};
		\node [style=none] (111) at (0.25, 2.45) {};
		\node [style=none] (112) at (-2.75, 1.95) {};
		\node [style=none] (113) at (-2, 1.95) {};
		\node [style=none] (114) at (-0.5, 1.95) {};
		\node [style=none] (115) at (0.25, 1.95) {};
		\node [style=none] (116) at (-1.25, 2.45) {};
		\node [style=none] (117) at (-1.25, 1.95) {};
		\node [style=none] (118) at (-1.25, 1.7) {Input Features};
		\node [style=none] (120) at (-1.625, 7.375) {};
		\node [style=none] (124) at (-1.625, 6.875) {};
		\node [style=none] (127) at (-0.875, 7.375) {};
		\node [style=none] (128) at (-0.875, 6.875) {};
		\node [style=none] (129) at (-1.25, 7.625) {Output Label};
		\node [style=none] (130) at (-1.25, 6.875) {};
	\end{pgfonlayer}
	\begin{pgfonlayer}{edgelayer}
		\draw (1.center)
			 to [bend left=90, looseness=1.25] (0.center)
			 to (2.center)
			 to [bend left=90, looseness=1.25] (3.center)
			 to cycle;
		\draw (33.center)
			 to [bend left=90, looseness=1.25] (32.center)
			 to (34.center)
			 to [bend left=90, looseness=1.25] (35.center)
			 to cycle;
		\draw [style=blackArrow] (30.center) to (37.center);
		\draw (41.center)
			 to [bend left=90, looseness=1.25] (40.center)
			 to (42.center)
			 to [bend left=90, looseness=1.25] (43.center)
			 to cycle;
		\draw [bend left=90, looseness=1.25] (67.center) to (66.center);
		\draw (66.center) to (68.center);
		\draw [bend left=90, looseness=1.25] (68.center) to (69.center);
		\draw (69.center) to (67.center);
		\draw [style=blackArrow] (64.center) to (71.center);
		\draw [style=blackArrow] (38.center) to (65.center);
		\draw (75.center)
			 to [bend left=90, looseness=1.25] (74.center)
			 to (76.center)
			 to [bend left=90, looseness=1.25] (77.center)
			 to cycle;
		\draw [style=blackArrow] (72.center) to (99.center);
		\draw [style=grey edge only] (108.center) to (112.center);
		\draw [style=grey edge only] (112.center) to (115.center);
		\draw [style=grey edge only] (115.center) to (111.center);
		\draw [style=grey edge only] (111.center) to (108.center);
		\draw [style=grey edge only] (109.center) to (113.center);
		\draw [style=grey edge only] (110.center) to (114.center);
		\draw [style=grey edge only] (116.center) to (117.center);
		\draw [style=blackArrow] (116.center) to (31.center);
		\draw [style=grey edge only] (120.center) to (124.center);
		\draw [style=grey edge only] (127.center) to (128.center);
		\draw [style=blackArrow] (98.center) to (130.center);
		\draw [style=grey edge only] (128.center) to (124.center);
		\draw [style=grey edge only] (120.center) to (127.center);
	\end{pgfonlayer}
\end{tikzpicture}}
            \caption{Neural Networks with Linear Mixed Effects (NN-LME)~\cite{xiong2019mixed}}
            \label{fig:nnlme}
        \end{subfigure}\hfill%
        \begin{subfigure}[t]{.24\textwidth}
            \resizebox{.9\textwidth}{!}{\begin{tikzpicture}
    \clip(-3.4,0.25) rectangle (0.8,7.9);
	\begin{pgfonlayer}{nodelayer}
		\node [style=none] (0) at (-2.9, 2.45) {};
		\node [style=none] (1) at (-2.9, 1.325) {};
		\node [style=none] (2) at (0.4, 2.45) {};
		\node [style=none] (3) at (0.4, 1.325) {};
		\node [style=circleB] (4) at (-2.75, 1.575) {};
		\node [style=circleB3] (8) at (-2.625, 2.175) {};
		\node [style=circleB2] (12) at (-2.75, 2.175) {};
		\node [style=circleB1] (13) at (-2.875, 2.175) {};
		\node [style=none] (14) at (-2.75, 1.875) {\small +};
		\node [style=circleB] (15) at (-1.75, 1.575) {};
		\node [style=circleB3] (16) at (-1.625, 2.175) {};
		\node [style=circleB2] (17) at (-1.75, 2.175) {};
		\node [style=circleB1] (18) at (-1.875, 2.175) {};
		\node [style=none] (19) at (-1.75, 1.875) {\small +};
		\node [style=circleB] (20) at (-0.75, 1.575) {};
		\node [style=circleB3] (21) at (-0.625, 2.175) {};
		\node [style=circleB2] (22) at (-0.75, 2.175) {};
		\node [style=circleB1] (23) at (-0.875, 2.175) {};
		\node [style=none] (24) at (-0.75, 1.875) {\small+};
		\node [style=circleB] (25) at (0.25, 1.575) {};
		\node [style=circleB3] (26) at (0.375, 2.175) {};
		\node [style=circleB2] (27) at (0.25, 2.175) {};
		\node [style=circleB1] (28) at (0.125, 2.175) {};
		\node [style=none] (29) at (0.25, 1.875) {\small +};
		\node [style=none] (30) at (-1.25, 2.45) {};
		\node [style=none] (31) at (-1.25, 1.325) {};
		\node [style=none] (32) at (-2.85, 3.05) {};
		\node [style=none] (33) at (-2.85, 2.575) {};
		\node [style=none] (34) at (0.45, 3.05) {};
		\node [style=none] (35) at (0.45, 2.575) {};
		\node [style=none] (36) at (-1.25, 2.575) {};
		\node [style=none] (37) at (-1.25, 2.575) {};
		\node [style=none] (38) at (-1.25, 3.05) {};
		\node [style=none] (39) at (-1.25, 2.8) {Nonlinear Layer};
		\node [style=none] (40) at (-2.9, 4.3) {};
		\node [style=none] (41) at (-2.9, 3.175) {};
		\node [style=none] (42) at (0.4, 4.3) {};
		\node [style=none] (43) at (0.4, 3.175) {};
		\node [style=circleB] (44) at (-2.75, 3.425) {};
		\node [style=circleB3] (45) at (-2.625, 4.025) {};
		\node [style=circleB2] (46) at (-2.75, 4.025) {};
		\node [style=circleB1] (47) at (-2.875, 4.025) {};
		\node [style=none] (48) at (-2.75, 3.725) {\small +};
		\node [style=circleB] (49) at (-1.75, 3.425) {};
		\node [style=circleB3] (50) at (-1.625, 4.025) {};
		\node [style=circleB2] (51) at (-1.75, 4.025) {};
		\node [style=circleB1] (52) at (-1.875, 4.025) {};
		\node [style=none] (53) at (-1.75, 3.725) {\small +};
		\node [style=circleB] (54) at (-0.75, 3.425) {};
		\node [style=circleB3] (55) at (-0.625, 4.025) {};
		\node [style=circleB2] (56) at (-0.75, 4.025) {};
		\node [style=circleB1] (57) at (-0.875, 4.025) {};
		\node [style=none] (58) at (-0.75, 3.725) {\small+};
		\node [style=circleB] (59) at (0.25, 3.425) {};
		\node [style=circleB3] (60) at (0.375, 4.025) {};
		\node [style=circleB2] (61) at (0.25, 4.025) {};
		\node [style=circleB1] (62) at (0.125, 4.025) {};
		\node [style=none] (63) at (0.25, 3.725) {\small +};
		\node [style=none] (64) at (-1.25, 4.3) {};
		\node [style=none] (65) at (-1.25, 3.175) {};
		\node [style=none] (66) at (-2.85, 4.9) {};
		\node [style=none] (67) at (-2.85, 4.425) {};
		\node [style=none] (68) at (0.45, 4.9) {};
		\node [style=none] (69) at (0.45, 4.425) {};
		\node [style=none] (70) at (-1.25, 4.425) {};
		\node [style=none] (71) at (-1.25, 4.425) {};
		\node [style=none] (72) at (-1.25, 4.9) {};
		\node [style=none] (73) at (-1.25, 4.65) {Nonlinear Layer};
		\node [style=none] (74) at (-2.9, 6.15) {};
		\node [style=none] (75) at (-2.9, 5.025) {};
		\node [style=none] (76) at (0.4, 6.15) {};
		\node [style=none] (77) at (0.4, 5.025) {};
		\node [style=circleB] (78) at (-2.75, 5.275) {};
		\node [style=circleB3] (79) at (-2.625, 5.875) {};
		\node [style=circleB2] (80) at (-2.75, 5.875) {};
		\node [style=circleB1] (81) at (-2.875, 5.875) {};
		\node [style=none] (82) at (-2.75, 5.575) {\small +};
		\node [style=circleB] (83) at (-1.75, 5.275) {};
		\node [style=circleB3] (84) at (-1.625, 5.875) {};
		\node [style=circleB2] (85) at (-1.75, 5.875) {};
		\node [style=circleB1] (86) at (-1.875, 5.875) {};
		\node [style=none] (87) at (-1.75, 5.575) {\small +};
		\node [style=circleB] (88) at (-0.75, 5.275) {};
		\node [style=circleB3] (89) at (-0.625, 5.875) {};
		\node [style=circleB2] (90) at (-0.75, 5.875) {};
		\node [style=circleB1] (91) at (-0.875, 5.875) {};
		\node [style=none] (92) at (-0.75, 5.575) {\small+};
		\node [style=circleB] (93) at (0.25, 5.275) {};
		\node [style=circleB3] (94) at (0.375, 5.875) {};
		\node [style=circleB2] (95) at (0.25, 5.875) {};
		\node [style=circleB1] (96) at (0.125, 5.875) {};
		\node [style=none] (97) at (0.25, 5.575) {\small +};
		\node [style=none] (98) at (-1.25, 6.15) {};
		\node [style=none] (99) at (-1.25, 5.025) {};
		\node [style=none] (100) at (-2.85, 6.75) {};
		\node [style=none] (101) at (-2.85, 6.275) {};
		\node [style=none] (102) at (0.45, 6.75) {};
		\node [style=none] (103) at (0.45, 6.275) {};
		\node [style=none] (104) at (-1.25, 6.275) {};
		\node [style=none] (105) at (-1.25, 6.275) {};
		\node [style=none] (106) at (-1.25, 6.75) {};
		\node [style=none] (107) at (-1.25, 6.5) {Nonlinear Layer};
		\node [style=none] (108) at (-2.75, 1.2) {};
		\node [style=none] (109) at (-2, 1.2) {};
		\node [style=none] (110) at (-0.5, 1.2) {};
		\node [style=none] (111) at (0.25, 1.2) {};
		\node [style=none] (112) at (-2.75, 0.7) {};
		\node [style=none] (113) at (-2, 0.7) {};
		\node [style=none] (114) at (-0.5, 0.7) {};
		\node [style=none] (115) at (0.25, 0.7) {};
		\node [style=none] (116) at (-1.25, 1.2) {};
		\node [style=none] (117) at (-1.25, 0.7) {};
		\node [style=none] (118) at (-1.25, 0.45) {Input Features};
		\node [style=none] (119) at (-2.375, 7.375) {};
		\node [style=none] (120) at (-1.625, 7.375) {};
		\node [style=none] (121) at (-0.125, 7.375) {};
		\node [style=none] (123) at (-2.375, 6.875) {};
		\node [style=none] (124) at (-1.625, 6.875) {};
		\node [style=none] (125) at (-0.125, 6.875) {};
		\node [style=none] (127) at (-0.875, 7.375) {};
		\node [style=none] (128) at (-0.875, 6.875) {};
		\node [style=none] (129) at (-1.25, 7.625) {Output Labels};
		\node [style=none] (130) at (-1.25, 6.875) {};
	\end{pgfonlayer}
	\begin{pgfonlayer}{edgelayer}
		\draw (1.center)
			 to [bend left=90, looseness=1.25] (0.center)
			 to (2.center)
			 to [bend left=90, looseness=1.25] (3.center)
			 to cycle;
		\draw (33.center)
			 to [bend left=90, looseness=1.25] (32.center)
			 to (34.center)
			 to [bend left=90, looseness=1.25] (35.center)
			 to cycle;
		\draw [style=blackArrow] (30.center) to (37.center);
		\draw (41.center)
			 to [bend left=90, looseness=1.25] (40.center)
			 to (42.center)
			 to [bend left=90, looseness=1.25] (43.center)
			 to cycle;
		\draw [bend left=90, looseness=1.25] (67.center) to (66.center);
		\draw (66.center) to (68.center);
		\draw [bend left=90, looseness=1.25] (68.center) to (69.center);
		\draw (69.center) to (67.center);
		\draw [style=blackArrow] (64.center) to (71.center);
		\draw [style=blackArrow] (38.center) to (65.center);
		\draw (75.center)
			 to [bend left=90, looseness=1.25] (74.center)
			 to (76.center)
			 to [bend left=90, looseness=1.25] (77.center)
			 to cycle;
		\draw [bend left=90, looseness=1.25] (101.center) to (100.center);
		\draw (100.center) to (102.center);
		\draw [bend left=90, looseness=1.25] (102.center) to (103.center);
		\draw (103.center) to (101.center);
		\draw [style=blackArrow] (98.center) to (105.center);
		\draw [style=blackArrow] (72.center) to (99.center);
		\draw [style=grey edge only] (108.center) to (112.center);
		\draw [style=grey edge only] (112.center) to (115.center);
		\draw [style=grey edge only] (115.center) to (111.center);
		\draw [style=grey edge only] (111.center) to (108.center);
		\draw [style=grey edge only] (109.center) to (113.center);
		\draw [style=grey edge only] (110.center) to (114.center);
		\draw [style=grey edge only] (116.center) to (117.center);
		\draw [style=blackArrow] (116.center) to (31.center);
		\draw [style=grey edge only] (119.center) to (123.center);
		\draw [style=grey edge only] (120.center) to (124.center);
		\draw [style=grey edge only] (121.center) to (125.center);
		\draw [style=grey edge only] (127.center) to (128.center);
		\draw [style=grey edge only] (119.center) to (121.center);
		\draw [style=grey edge only] (123.center) to (125.center);
		\draw [style=blackArrow] (106.center) to (130.center);
	\end{pgfonlayer}
\end{tikzpicture}}
            \caption{Neural Mixed Effects (NME), our approach}
            \label{fig:nme}
        \end{subfigure}
    
    \caption{Visual comparison of our approach, Neural mixed Effects (NME), and previous approaches. NME enables person-specific parameters at any layer to represent nonlinear person-specific trends. Person-generic ($\bar{\theta}$) and person-specific ($\theta^i$) parameters are combined by summing, i.e., $\bar{\theta} + \theta^i$.}
    \Description{Structural differences between different mixed effect models.}
    \label{fig:mixed}
\end{figure*}
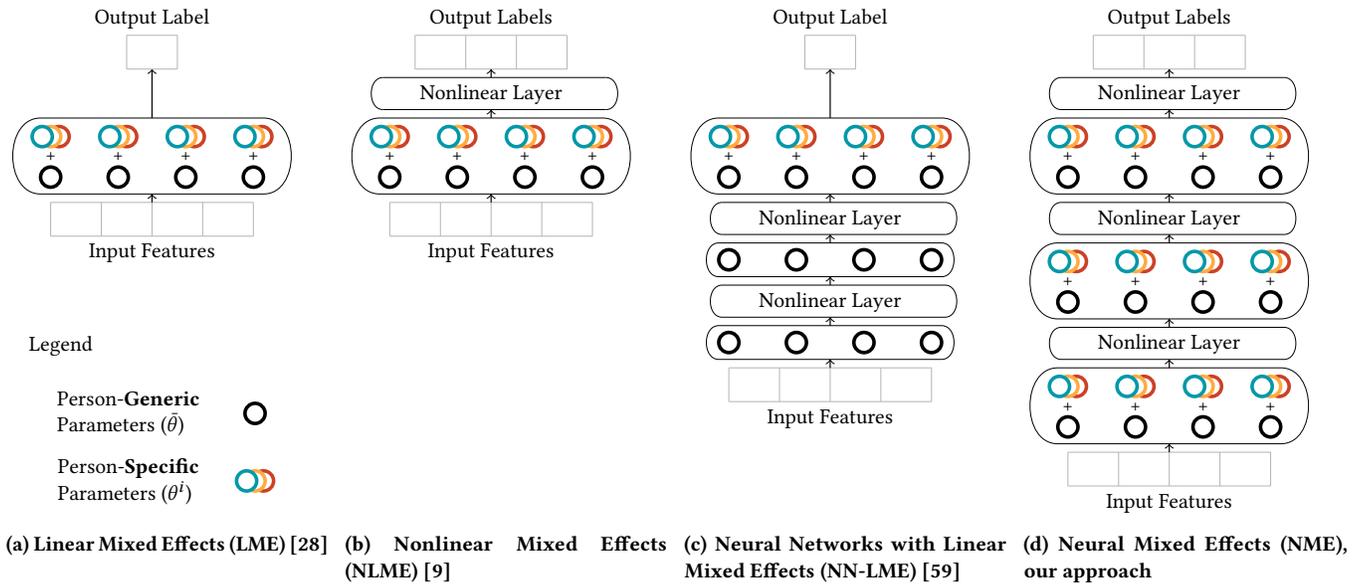

Personalized prediction is a machine learning approach that predicts a person's future observations based on their past labeled observations. This type of model is typically used for sequential tasks that would be difficult without knowledge of the person, such as predicting daily mood from only smartphone data or predicting affective state sequences where transitions between states might be influenced by depression~\cite{pratap2019accuracy,pedrelli2020monitoring}. As illustrated in \autoref{fig:problem}, a personalized model benefits by combining two types of trends (a) person-generic trends shared across people, such as being happier on weekends, and (b) unique person-specific trends, such as stressful weekly meetings or weekly socializing with friends. Person-specific trends can be challenging for machine learning models, even when trained on data from these people, as they might average out across people: as exemplified in  \autoref{fig:problem} when the more positive mood from a person's socializing coincides with the more negative mood of another person's stressful meeting.

Mixed effect models\footnote{In statistics, the person-generic trends are often referred to as \emph{fixed effects} and the person-specific trends as \emph{random effects}. The name mixed effects comes from mixing both fixed and random effects.} are popular in statistics to study person-generic and person-specific trends by combining person-generic and person-specific parameters~\cite{laird1982random}. Linear mixed effect (LME) models have recently been gaining popularity in machine learning for personalizing models~\cite{tandon2006neural,tran2017random,xiong2019mixed,ngufor2019mixed,mandel2021neural,levy2021mixed,anonymous2022mixed,lewis2023mixed,simchoni2022integrating,linkedin}. Integrating LME with neural networks is currently limited to linear person-specific trends: person-specific parameters can only be in the last linear layer of a neural network as illustrated in \autoref{fig:nnlme}. This rules out person-specific parameters in the remaining layers, i.e., nonlinear person-specific parameters. Separately from work with neural networks,  nonlinear mixed effect approaches were proposed, but their optimization does not scale to large neural networks with many layers and parameters~\cite{saem}.

In this paper, we propose Neural Mixed Effect (NME) models to learn nonlinear person-specific parameters in a scalable manner. Our NME models combine the efficient optimization of neural networks with the person-specific parameters of nonlinear mixed effect models. NME learns nonlinear person-specific parameters by enabling them anywhere in a  nonlinear neural network, as shown in \autoref{fig:nme}. We demonstrate integrating our NME approach into two model architectures. We evaluate performance primarily on Multi-Layer Perceptrons (MLPs) for better comparison with previous MLP-LME work. To demonstrate NME for more complex models that yet have some interpretable parameters, we integrate NME with neural Conditional Random Fields (CRFs) to classify states in a temporal sequence~\cite{durrett2015neural}. CRFs explicitly model a sequence's temporal dynamics and allow us to interpret the person-specific temporal transitions between states.

We evaluate NME on six unimodal and multimodal datasets, including a smartphone dataset to predict daily mood and a mother-adolescent dataset to predict affective state sequences where half the mothers experience symptoms of depression. We analyze the interpretable person-specific transition parameters in the CRF and hypothesize that they differ between families where mothers experience symptoms of depression.

\begin{table*}
    \centering
    \begin{tabular}{lcccc}
        \toprule
         & Linear Mixed Effects & Nonlinear Mixed Effects & Neural Networks with & Neural Mixed Effects \\
         & (LME) & (NLME) & Linear Mixed Effects (NN-LME) & (NME)\\
        \midrule
        Nonlinear Model & \xmark & \cmark &\cmark* & \cmark\\
        Dataset Scalability & \xmark & \cmark & \xmark & \cmark\\
        Model Scalability & \cmark & \xmark & \cmark & \cmark\\
        \bottomrule
    \end{tabular}
    \caption{Comparison of NME with previous approaches. LME models do not scale well with too many observations per person. The sampling-based optimization of NLME does not scale well with too many parameters. NN-LME has nonlinear person-generic parameters, but it re-use the optimization of LME, which (*) limits NN-LME to linear person-specific parameters and it does not scale as well for large datasets. Our proposed NME combines the efficient optimization of neural networks with the nonlinear persons-specific parameters of mixed effect models.}
    \label{tab:optimization}
\end{table*}

\section{Technical and Related Background}
Mixed effect models were proposed in statistics for data that is not independent and identically distributed, e.g., longitudinal data from multiple people~\cite{laird1982random}. In statistics, the goal of mixed effect models is often to study research questions about person-generic trends, referred to as fixed effects, and person-specific trends, referred to as random effects. Mixed effect models include a penalty term to regularize the person-specific parameters (denoted as $\bm{\theta}^i$) so that they learn only what the person-generic parameters (denoted as $\bar{\bm{\theta}}$) cannot learn. The technical challenge when optimizing mixed effect models is to separate fixed and random effects since they affect each other, e.g., a random bias term can affect the fixed slope of linear mixed effect models~\cite{simpson1951interpretation}.

We briefly highlight the optimization of linear and nonlinear mixed effect models, review related work that explored combinations of neural networks and mixed effect models, and then contrast mixed models with multitask learning.

\textbf{Linear Mixed Effects (LME):} For an observation from the $i$-th person represented by a feature vector $\bm{X}$, a linear mixed effects model infers the prediction as $\hat{y} = (\bar{\bm{\theta}} + \bm{\theta}^i)^T \bm{X}$, see \autoref{fig:lme}. For efficient optimization, it is often assumed that the random effects $\bm{\theta}^i$ follow a multivariate normal distribution with zero mean and covariance $\bm{\Sigma}$. A popular method to optimize LME models is an Expectation-Maximization (EM) algorithm that minimizes the mean squared error~\cite{lindstrom1988newton}. The challenging part of this EM algorithm is that a matrix needs to be inverted for each person $i$, where the matrix size is the number of observations for person $i$. This makes it challenging to optimize LME models when a person has many observations, i.e., LME models do not easily scale to large datasets.

\textbf{Nonlinear Mixed Effects (NLME):} Nonlinear mixed effect models are used to model nonlinear person-specific trends, for example, in pharmacometrics~\cite{owen2014introduction}. As shown in \autoref{fig:nlme}, random effects can be anywhere in a nonlinear model $\hat{y} = f(\bm{X}; \bar{\bm{\theta}} + \bm{\theta}^i)$ making their optimization more challenging. While multiple optimization approaches exist for nonlinear mixed effects~\cite{lindstrom1990nonlinear,pinheiro1995approximations,saem,bates2014fitting}, most modern nonlinear mixed effect approaches find an approximate solution using random walk Metropolis sampling~\cite{saem,karimi2020f}. One downside of this sampling approach is that it converges slowly for large models with many parameters~\cite{karimi2020f}. One upside, compared to LME, is that this sampling approach scales well with many observations as it does not require matrix inversions that depend on the number of people or observations. 

\textbf{Neural Networks with Linear Mixed Effects (NN-LME):} 
LME models have been combined with neural networks to improve performance for tasks involving longitudinal data from multiple people, such as for mood and mental health-related tasks~\cite{tandon2006neural,tran2017random,xiong2019mixed,mandel2021neural,anonymous2022mixed,simchoni2022integrating,linkedin}. All of these combinations follow the same mathematical formulation of $\hat{y} = (\bar{\bm{\theta}} + \bm{\theta}^i)^T f(\bm{X}; \bm{\theta}_\textrm{neural})$, see \autoref{fig:nnlme}, where $\bm{\theta}_\textrm{neural}$ are the person-generic parameters of the neural network. These combinations can be seen as simply placing an LME model on top of a neural network. Most NN-LME approaches use the same EM algorithm as LME models~\cite{lindstrom1988newton}. The only difference is that the neural network parameters $\bm{\theta}_\textrm{neural}$ become part of the fixed effects, meaning the neural network needs to be trained until convergence within every E-step, which can be slow for large neural networks. By re-using the same EM algorithm from LME models, its limitations apply: the random effects will minimize the mean squared error and NN-LME will not easily scale to large datasets. While two approaches extend beyond the means squared error by finding an approximate solution for binary classification\cite{simchoni2022integrating,linkedin}, their work does not generalize to multiclass classification.

Our proposed Neural Mixed Effects (NME) approach is a significant generalization of previous work by allowing person-specific parameters, i.e., random effects, anywhere in neural networks where even the last layer can be nonlinear. Our proposed NME model is also scalable to large datasets and large models by efficiently optimizing the NLME objective with stochastic gradient descent. We summarize this comparison in \autoref{fig:mixed} and \autoref{tab:optimization}

\textbf{Multitask Models:} 
Assuming not all model parameters have a person-specific component, mixed models are similar to multitask models where each task corresponds to a person~\cite{caruana1997multitask,song2022learning}. The two main differences are 1) mixed models have a person-generic ("shared") component even for parameters that have a person-specific component and 2) while multitask models can have an additional explicit regularization between the task-specific parameters~\cite{evgeniou2004regularized,taylor2017personalized}, mixed models do not require a hyper-parameter to determine the strength of this regularization as $\bm{\Sigma}$ is learned.

\section{Problem Statement}
Our main goal is personalized prediction: predicting a person's future observations by training on their past observations. The problem of personalized prediction using mixed effects can be formalized as follows. Given a training dataset with $n$ people and $n_i$ observations for the $i$-th person $\{(\bm{X}^i_j, y^i_j)~|~i \in [1, n],~j \in [1, n_i]\}$ and a test dataset with unseen observations from the same people, the goal is to learn a function $f(\bm{X}^i_j; \bm{\theta})$ predicting $y^i_j$ where the parameters $\bm{\theta}$ are expressed as the sum of a person-generic $\bar{\bm{\theta}}$ and a person-specific component $\bm{\theta}^i$.

\section{Neural Mixed Effect Models} \label{sec:nme}
Mixed effect models are gaining popularity in machine learning for personalized predictions as they combine person-generic and person-specific parameters. In this section, we present our generalization named Neural Mixed Effects (NME) model to better integrate mixed effect models in neural networks through a more scalable optimization and by allowing person-specific parameters anywhere. The advantage of our proposed NME approach is that it enables any neural network architecture to have person-specific parameters $\bm{\theta}^i$ as long as its original parameters (which we will refer to as person-generic parameters $\bar{\bm{\theta}}$) can be optimized with gradient descent. The only difference is that the person-specific components $\bm{\theta}^i$ also need to be stored and optimized. When making predictions for person $i$, the neural network parameters become the sum of these two components $\bar{\bm{\theta}} + \bm{\theta}^i$. Similar to multitask learning, not all parameters need a person-specific component. If parameters have no person-specific components, the parameters are equal to the person-generic components $\bar{\bm{\theta}}$.

We first focus on the optimization process in \autoref{ssec:optimization}, then show that NME is a nonlinear mixed effects model in \autoref{ssec:mixed}, and finally, we describe in \autoref{ssec:crf} how to predict sequences using a neural Conditional Random Field (CRF) and how we combine it with NME.

\subsection{Optimization}\label{ssec:optimization}

The goal is to learn person-specific parameters $\bm{\theta}^i$ representing person-specific trends, i.e., that cannot be learned by the person-generic parameters $\bar{\bm{\theta}}$. In addition to minimizing a downstream loss function $l$, mixed effect models separate person-generic and person-specific trends by regularizing the person-specific parameters. This regularizing encourages the person-specific parameters $\bm{\theta}^i$ to only focus on what cannot be learned by the unregularized person-generic parameters $\bar{\bm{\theta}}$. Following previous NN-LME work, we regularized the person-specific parameters by assuming that they follow a multivariate normal distribution with zero mean and covariance matrix $\bm{\Sigma} \in \mathbb{R}^{\textrm{dim}(\bm{\theta}^i) \times\textrm{dim}(\bm{\theta}^i)}$, where $\textrm{dim}(\bm{\theta}^i)$ is the number of person-specific parameters. $\bm{\Sigma}$ is the same for all people. To make the regularization invariant to the scale of different downstream loss functions, mixed effect models have, next to  $\bm{\Sigma}$, a second weighting factor $\sigma^2$ that represents the average downstream loss. The resulting loss function of NME is  

\begin{align}
    \sum_{i=1}^{n} \left[ \frac{1}{\sigma^2}\sum_{j=1}^{n_i} l(y^i_j, f(\bm{X}^i_j; \bar{\bm{\theta}} + \bm{\theta}^i)) \right] +  \bm{\theta}^{iT}\bm{\Sigma}^{-1} \bm{\theta}^i~.  \label{eq:min}
\end{align}

The left term of \autoref{eq:min} optimizes $\bar{\bm{\theta}} + \bm{\theta}^i$ for best downstream performance while the right term regularizes the person-specific parameters $\bm{\theta}^i$. As we have separate persons-specific parameters $\bm{\theta}^i$  for each person $i$ but apply the same regularization, we are likely to learn larger person-specific parameters when a person has many observations: as the left term, the sum over the number of observations for a person is more likely to outweigh the regularization term on the right when a person has many observations. Intuitively, this improves performance the most when we have many observations for a person and helps prevent overfitting for a person with only a few observations.

Optimization of \autoref{eq:min} is performed with stochastic gradient descent in batches, where the regularization term on the right is scaled by how many observations a person has in the current batch $B$. The right part of \autoref{eq:min} becomes

% \bm{1} for arxiv, \mathds{1} for everyone else
\begin{align}
    \frac{\sum_{\bm{X}^k_j \in B} \mathds{1}(k=i)}{n_i}\bm{\theta}^{iT}\bm{\Sigma}^{-1} \bm{\theta}^i \label{eq:bmin}
\end{align}

where the indicator function $\mathds{1}(k=i)$ is $1$ when the observation $\bm{X}^k_j$ is from the $i$-th person, i.e., $k=i$.

After each epoch of minimizing \autoref{eq:min}, we update $\sigma^2$ to the new average downstream loss $l$ of the training set and $\bm{\Sigma}$ to the sample covariance matrix of the person-specific parameters $\bm{\theta}^i$.

Fortunately, it is common in mixed effect modeling to assume that the person-specific parameters are independent of each other~\cite{wolfinger1993covariance,saem}, which reduces $\bm{\Sigma}$ to an easy-to-invert diagonal matrix. This allows us to efficiently optimize \autoref{eq:min} even for large models with many person-specific parameters. NMEs with this assumption are as fast as multitask models when having the same person/task-specific parameters. As seen from \autoref{eq:min}, the NME objective scales linearly with the number of people and their observations enabling NME to scale to even large datasets.

To summarize, 1) NME allows person-specific parameters anywhere in a neural network, 2) NME uses stochastic gradient descent to optimize even large models with many person-specific parameters efficiently, and 3) NME scales linearly with the dataset size.

\begin{figure}
    \centering
    \resizebox{.9\linewidth}{!}{\input{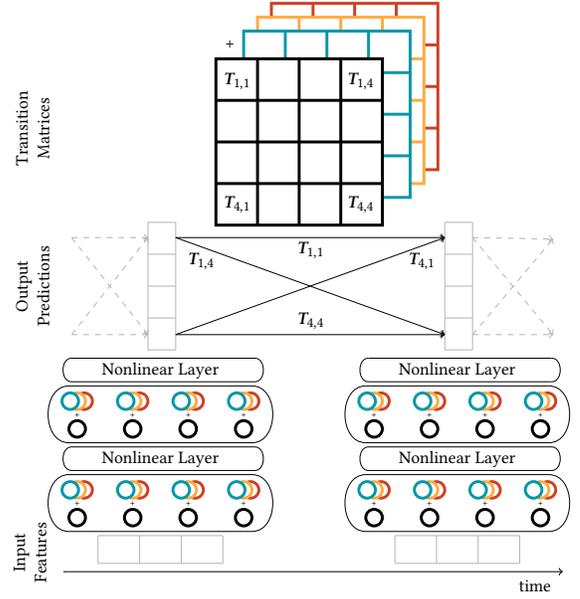}}
    \caption{Illustration of the NME-CRF with person-specific parameters everywhere. An MLP predicts the initial output predictions which are refined by the CRF using the transition matrix $\bm{T}$.}
    \label{fig:crf}
    \Description{Diagram of the CRF that combines the transition matrix and the output from an MLP in the viterbi algorithm to make a prediction.}
\end{figure}

\begin{table*}
    \caption{Dataset characteristics. With the \emph{calendar} modality we refer to metadata including the year and the weekday.}
    \center
    \begin{tabular}{rlcS[table-format=3.0]S[table-format=5.0]l}
        \toprule
        Dataset & Tasks & Group & {\#Groups} & {\#Observations} & Modalities\\
        \midrule
        Imdb~\cite{imdb} & Movie rating (regression) & Genre & 383 & 83143 & text\\
        News~\cite{news} & Number of shares on Facebook (regression)& Outlet & 598 & 60080 & calendar, text\\
        Spotify~\cite{tidytuesday} & Danceability rating (regression)& Genre & 58 & 26844 & acoustic, calendar, text\\
        IEMOCAP~\cite{busso2008iemocap} & Arousal and valence ratings (regression) & Person & 10 & 4784 & acoustic, text, vision\\ 
        MAPS~\cite{maps} & Daily self-assessed mood ratings (regression) & Person & 38 & 2122 & calendar, GPS, text, typing\\
        TPOT~\cite{wortwein2021human} & Four affective states (multiclass classification) & Person &195 & 15228 & acoustic, text, vision\\
        \bottomrule
    \end{tabular}
    \label{tab:datasets}
\end{table*}

\subsection{NME as a Nonlinear Mixed Effects Model}\label{ssec:mixed}

NME learns a nonlinear mixed effects model because its optimization procedure follows that of the nonlinear mixed effects solver saemix~\cite{saem}. saemix is designed to optimize nonlinear mixed effect models in statistics using random walk Metropolis sampling. However, sampling many parameters for neural networks is typically computationally challenging, converges slowly, and might lead to sub-optimal solutions~\cite{karimi2020f,papamarkou2022challenges,de2000sequential}. NME replaces sampling with gradient descent to scale to large neural networks with many person-specific parameters.

saemix is an approximation EM algorithm~\cite{delyon1999convergence}, which means the expectation step (E-step) is not required to have converged before continuing with the maximization step (M-step). When assuming that the person-specific parameters $\bm{\theta}^i$ follow a multivariate normal distribution with zero mean and covariance matrix $\bm{\Sigma}$, saemix incrementally minimizes \autoref{eq:min} during the E-step. During the M-step, saemix updates $\sigma^2$ and $\bm{\Sigma}$. Under general assumptions\footnote{Assuming $l(y^i_j, f(\bm{X}^i_j; \bar{\bm{\theta}} + \bm{\theta}^i)$ are conditionally independent given the person $i$ and follow a distribution in the exponential family.}, saemix will converge to a mixed effects model. NME reduces \autoref{eq:min} during each epoch, corresponding to the E-step. Updating $\sigma^2$ and $\bm{\Sigma}$ between epochs corresponds to the M-steps. As NME follows the optimization procedure of saemix, NME will also converge to a nonlinear mixed effects model.

\subsection{NME Conditional Random Fields}\label{ssec:crf}
When predicting states that have a temporal order, such as the sequence of affective states on the mother-adolescent dataset, it can be beneficial to account for temporal dynamics, e.g., how likely it is to transition from one state to the next. Accounting for temporal dynamics may not only improve performance, but it may also be possible to interpret which transition the model infers as more or less likely. If we can further learn person-specific transitions, we can interpret whether they differ, for example, between families where mothers experience symptoms of depression.

Conditional Random Fields (CRFs) are graphical models that can learn state transitions in an interpretable manner~\cite{lafferty2001conditional}. When the transitions are assumed to be time-invariant, i.e., they are constant across time, we can represent all possible transitions from one to the next state through one matrix $\bm{T} \in \mathbb{R}^{|\textrm{states}| \times |\textrm{states}|}$ where $|\textrm{states}|$ is the number of states. CRFs learn such a transition matrix $\bm{T}$. While CRFs have been combined with neural networks~\cite{durrett2015neural}, they have not been explored with person-specific parameters, as done in the NME approach. With our NME-CRF, we can learn person-specific transition matrices $\bm{T}=\bar{\bm{T}}+\bm{T}^i$, which allows us to analyze them.

Besides a transition matrix $\bm{T}$, a CRF needs to know how likely each state is at time $t$, which we infer using an MLP.  \autoref{fig:crf} provides an illustration of NME-CRF. The CRF model can be optimized using gradient descent by minimizing the following loss function
\begin{align}
    -\frac{
        \textrm{exp}\left(\sum_t^L f(\bm{X}^i_t; \bar{\bm{\theta}}+\bm{\theta}^i)) + (\bar{\bm{T}}+\bm{T}^i)_{y_{t-1}, y_t} \right)
    }{
        Z([\bm{X}^i_1, \dots{}, \bm{X}^i_L])
    } \label{eq:forwardbackward}
\end{align}
where $Z$ is a normalization function. We use the forward-backward algorithm to efficiently calculate \autoref{eq:forwardbackward}~\cite{binder1997space}. To combine the CRF with NME, \autoref{eq:forwardbackward} becomes the downstream loss $l$ in \autoref{eq:min}. At inference time, we use the viterbi algorithm to efficiently determine the most likely state sequence~\cite{binder1997space}.

\section{Experimental Setup}

We evaluate our NME approach on six unimodal and multimodal datasets, including both regression and multiclass classification tasks. For better comparison with previous approaches, we primarily integrate NME with MLPs. The mother-adolescent dataset has temporal state sequences allowing us to evaluate the NME-CRF. We perform a more detailed analysis of the learned parameters of the NME-CRF since it learns interpretable state transitions.

\subsection{Datasets}
We conduct experiments on six datasets, summarized in \autoref{tab:datasets}.

\textbf{Imdb~\cite{imdb}, News~\cite{news}, Spotify~\cite{tidytuesday}:} These are three public datasets used by previous NN-LME work~\cite{simchoni2022integrating}. We follow their experimental protocol and use the same features and labels. Instead of people being the grouping variable on these datasets, we have genres on Imdb and Spotify and outlets on the News datasets as a grouping variable, i.e., we learn genre-specific and outlet-specific parameters. Following previous work, we report the root mean squared error (RMSE) for these three datasets. For easier comparison across the three datasets, we normalize the RMSE by the standard deviation of the ground truth labels on the test set (NRMSE).

\begin{table*}
    \caption{Performance on six datasets with person-specific parameters in the last and all layers of the MLP. Best overall performance is underlined while best performance for the last/all layers is in bold. When a baseline is significantly worse than NME-MLP with person-specific parameters in the last or all layers, $L$ or $A$ are in superscript.}
    \center
    \begin{tabular}{clS[table-format = 1.3]S[table-format = 1.3]S[table-format = 1.3]S[table-format = 1.3]S[table-format = 1.3]S[table-format = 1.3]S[table-format = 1.3]}
        \toprule
        && {Imdb}	& {News} & {Spotify} & {IEMOCAP-A} & {IEMOCAP-V} & {MAPS} & {TPOT} \\
        \cmidrule(lr){3-3} \cmidrule(lr){4-4} \cmidrule(lr){5-5} \cmidrule(lr){6-6} \cmidrule(lr){7-7} \cmidrule(lr){8-8} \cmidrule(lr){9-9}
        && {NRMSE $\downarrow$} & {NRMSE $\downarrow$} & {NRMSE $\downarrow$} & {CCC $\uparrow$} & {CCC $\uparrow$} & {Pearon's $r$ $\uparrow$} & {Krippendorff $\alpha$ $\uparrow$} \\
        \midrule
        &Generic-MLP & 0.927$^{LA}$ & 0.841$^{LA}$& 0.711$^L$&0.510$^A$ &0.518$^A$ &0.119 & 0.355 \\
        \midrule
        \multirow{3}{*}{\rotatebox[origin=c]{90}{Last}} &MLP-LME~\cite{xiong2019mixed}  & 0.881$^L$ &0.630 & 0.685& 0.455$^L$& 0.466$^L$&0.143 & {---} \\ 
        &Specific-MLP & 0.891$^L$ &0.646$^L$ &0.794$^L$& 0.431$^L$& 0.354$^L$& 0.074& 0.347\\
        &NME-MLP (ours) & \underline{\textbf{0.846}} & \underline{\textbf{0.627}} & \underline{\textbf{0.679}}& \textbf{0.510} & \textbf{0.555}& \underline{\textbf{0.209}}& \underline{\textbf{0.367}}\\
        \midrule
        \multirow{2}{*}{\rotatebox[origin=c]{90}{All}}&Specific-MLP & 0.886$^A$ & 0.654$^A$ & 0.770$^A$ & 0.452$^A$ & 0.443$^A$ & 0.124 & 0.288$^A$\\
        &NME-MLP (ours)& \textbf{0.856} & \textbf{0.629} & \textbf{0.690} &\underline{\textbf{0.558}} & \underline{\textbf{0.559}} & \textbf{0.138} & \underline{\textbf{0.367}} \\
        \bottomrule
    \end{tabular}
    \label{tab:overview}
\end{table*}

\textbf{IEMOCAP~\cite{busso2008iemocap}:} The IEMOCAP dataset~\cite{busso2008iemocap} consists of dyadic interactions of five pairs of people, a total of ten people. Each pair is asked to improvise a set of emotionally charged interactions spontaneously. We separately predict arousal and valence ratings for each person on short utterances using features extracted by previous work~\cite{wortwein-etal-2022-beyond}, which includes statistics aggregated at the utterance-level of OpenFace 2.0~\cite{baltrusaitis2018openface}, openSMILE's eGeMaPs~\cite{egemaps}, and RoBERTa~\cite{liu2020roberta}. As is common for IEMOCAP, we use the concordance correlation coefficient (CCC)~\cite{lawrence1989concordance} as the evaluation metrics.

\textbf{MAPS~\cite{maps}}: Mobile Assessment for the Prediction of Suicide (MAPS) is a longitudinal dataset of smartphone data of adolescents with daily mood self-assessments~\cite{maps}. We predict the daily mood self-assessments using their phone activity from the past 24h. Inspired by previous phone-based mood prediction work~\cite{pratap2019accuracy,jacobson2020passive,liang-etal-2021-learning,auerbach2022geolocation}, we extracted the following features: LIWC dimensions~\cite{pennebaker2015development} and sentiment from Vader~\cite{hutto2014vader} of the typed text, the number of words, total time typing, the mean and variance of the typing speed, the weekday, the number of visited places based on GPS data as well as distance traveled and the average walking speed. The evaluation metric is Pearson's correlation coefficient $r$, which is well suited for evaluating how much of the mood variation we can predict.

\textbf{TPOT~\cite{nelson2021psychobiological}}: The Transitions in Parenting of Teens (TPOT) dataset contains video recordings of dyadic interactions between mothers and their adolescents~\cite{nelson2021psychobiological}. By design, mothers of half the dyads exhibit at least moderate depression symptoms at recruitment time and further had a treatment history for depression (referred to as the depressed group). The other half of mothers exhibits at most low symptoms, do not have a treatment history of depression, and had further no mental health treatment a month before recruitment (referred to as the non-depressed group). The interactions are typically 15 minutes long and focus on resolving areas of disagreement, such as participation in household chores. These interactions are annotated for each person for a sequence of four affective states (\emph{other}, \emph{aggressive}, \emph{dysphoric}, and \emph{positive}). These affective states are closely related to Living in Familial Environments codes~\cite{hops1995methodological,schwartz2014parenting}. The affective state annotations are onset annotations, i.e., a state is annotated when enough evidence is available to determine the affective state and last until enough evidence is available for the next onset. This annotation approach means that two consecutive segments will not have the same label, e.g., \emph{positive} will not follow \emph{positive}. When using the NME-MLP, we predict these segments independently of each other. As the NME-CRF allows us to model temporal dynamics, we jointly predict each person's sequence of segments. In both cases, we use the same features from previous work~\cite{wortwein2021human}, which are similar to the features on IEMOCAP but uses LIWC~\cite{pennebaker2015development} instead of RoBERTa. Following previous work, we report Krippendorff's $\alpha$ between the ground truth and the predicted labels.

\subsection{NME Models and Baselines}

Similar to previous work, we evaluate NME primarily in the context of MLPs (referred to as \textbf{NME-MLP}). Additionally, we evaluate NME using neural CRFs for the sequence prediction task on TPOT  (referred to as \textbf{NME-CRF}). Since our NME approach allows person-specific parameters anywhere in the model, we explore three approaches: 1) having person-specific parameters in only the last layer (denoted as \textbf{last}), 2) for the CRF to additionally have person-specific parameters in its transition matrix $\bm{T}$ (denoted as \textbf{last+$\bm{T}$}), and 3) having them everywhere in the model (denoted as \textbf{all}). \autoref{fig:crf} depicts the NME-CRF with person-specific parameters everywhere, including the transition matrix $\bm{T}$. 

We compare NME-MLP and NME-CRF to three baselines.

\textbf{Generic-MLP:} Generic-MLP is either an MLP or a CRF (\textbf{Generic-CRF}) with only person-generic parameters, i.e., $\bm{\theta} = \bar{\bm{\theta}}$. Generic-MLP corresponds to a conventional  MLP that is directly optimized with the downstream loss function $l$.

\textbf{Specific-MLP:} Specific-MLP is either an MLP or a CRF (\textbf{Specific-CRF}) with only person-specific parameters, i.e., $\bm{\theta} = \bm{\theta}^i$. The person-specific parameters are optimized with the downstream loss function $l$, i.e., they do not follow the NME approach. When evaluating person-specific parameters in only the last layer, we use person-generic parameters in all the previous layers of the MLP, i.e., $\bm{\theta} = \bar{\bm{\theta}}$ (the same as multitask learning with a task-specific last layer).

\textbf{MLP-LME~\cite{xiong2019mixed}:} Almost all previous MLP-LME work~\cite{tandon2006neural,tran2017random,xiong2019mixed,mandel2021neural} is based on the same EM algorithm~\cite{lindstrom1988newton}. We implement MLP-LME as described in previous work~\cite{xiong2019mixed}, which makes MLP-LME a baseline for regression tasks with person-generic and person-specific parameters in the last layer, i.e., $\bm{\theta} = \bar{\bm{\theta}} + \bm{\theta}^i$. MLP-LME has so far not been extended to multiclass classification, so we cannot evaluate MLP-LME on TPOT.

\subsection{Experimental Details}

For all datasets we have a within-person split of 60\% training, 20\% validation, and 20\% testing. For IEMOCAP, MAPS, and TPOT, the first 60\% of the observations per person are used for training, the following observations for validation, and the last observations for testing. This is done to avoid temporally correlated observations that would invalidate the validation or test set.

All models are implemented in PyTorch~\cite{Paszke_PyTorch_An_Imperative_2019} and optimized with Adam~\cite{kingma2014adam}. Their hyper-parameter are determined using a gridsearch which includes the learning rate, the number of layers in the MLP and their width, and L2 weight decay. Model validation is based on the validation set performance. All models are trained on consumer-level graphic cards, such as, the NVidia RTX 3080 Ti.

All input features are z-normalized on the training set. For regression tasks, the ground truth is also z-normalized based on the training set. The mean squared error is the loss function $l$ for all regression tasks. For the MLP on TPOT, we minimize the cross entropy loss, while the forward-backward algorithm is used for the CRF on TPOT to minimize \autoref{eq:forwardbackward}. Features from different modalities are combined through early fusion. 

When reporting performance metrics, we first calculate them within each person and then report the average. This allows us to focus on the within-person performance and avoids Simpson's paradox~\cite{simpson1951interpretation}. Significance tests are conducted with paired person-clustered bootstrapping~\cite{ren2010nonparametric} using $p=0.05$ and 10,000 resamplings at the person-level\footnote{For each person, calculate the performance metric and take their difference between two models. Then bootstrap the differences by resampling 10,000 times with replacement to derive 95\% confidence intervals using percentiles.}. To determine the performance metrics reliably, we need a large enough test set per person: we remove people from all experiments if we have less than ten observations from them.

\section{Results and Discussion}

\begin{figure}
    \centering
    \resizebox{.8\linewidth}{!}{\input{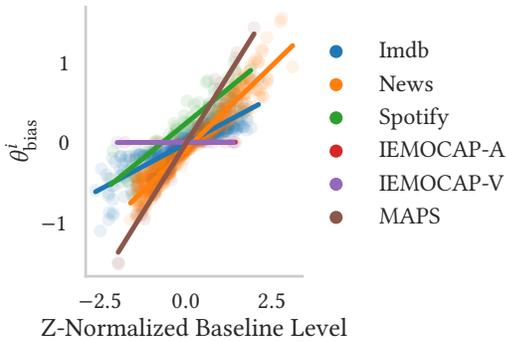}}
    \caption{Correlation between the baseline level (ground truth on the training set) and the last bias term $\theta^i_\textrm{bias}$ of NME-MLP.}
    \Description{Scatterplot between the the baseline level of each person and MLP-NME's last bias term.}
    \label{fig:bias}
\end{figure}

We first present the NME-MLP experiments across all six datasets and then focus on analyzing the NME-CRF multiclass classification experiments on the TPOT dataset.%, which incorporates the temporal information of the sequence of affective states.

\subsection{NME-MLP Experiments}
\textbf{Last layer with person-specific parameters:} We first evaluate NME-MLP with person-specific parameters in only the last layer for a direct comparison with MLP-LME~\cite{xiong2019mixed}. NME-MLP performs numerically equal or better than all three baselines (Generic-MLP, Specific-MLP, and MLP-LME) on the six datasets, see the top half of \autoref{tab:overview}. While Specific-MLP incurs a performance drop for the two smaller datasets, i.e., IEMOCAP and MAPS, NME-MLP maintains or improves performance indicating that it is important to have both person-generic and person-specific parameters. Unlike current MLP-LME implementations, NME-MLP can also be applied to multiclass classification on the TPOT dataset. NME-MLP again performs numerically better than its baselines. As indicated by the superscripts in \autoref{tab:overview}, NME performs in many cases statistically significantly better compared to its baselines.

\textbf{All layers with person-specific parameters:} As illustrated in \autoref{fig:nme}, NME enables person-specific parameters anywhere in a neural network. The bottom half of \autoref{tab:overview} summarizes the performance with person-specific parameters everywhere. NME-MLP numerically outperforms Specific-MLP and Generic-MLP. Having person-specific parameters everywhere also leads to the best performance across all IEMOCAP experiments suggesting that people in IEMOCAP may have nonlinear person-specific trends. 

\begin{table}
    \caption{Performance of the CRF on TPOT. Best overall performance is underlined while best performance for the last/all layers is in bold.}
    \center
    \begin{tabular}{clS[table-format = 1.3]}
        \toprule
        && {Krippendorff $\alpha$ $\uparrow$} \\
        \midrule
        &Generic-CRF & 0.467\\
        \midrule
        \multirow{2}{*}{\rotatebox[origin=c]{90}{\parbox{.6cm}{Last + $\bm{T}$}}} &Specific-CRF & 0.485\\
        & NME-CRF (ours) & \underline{\textbf{0.494}}\\
        \midrule
        \multirow{2}{*}{\rotatebox[origin=c]{90}{All}} &Specific-CRF  & 0.317$^A$ \\
        &NME-CRF (ours) & \textbf{0.470}\\
        \bottomrule
    \end{tabular}
    \label{tab:crf}
\end{table}

\textbf{Interpretation of baseline levels:} NME-MLPs for regression infer their prediction as $\hat{y} = (\bar{\bm{\theta}}+\bm{\theta}^i)^T \bm{Z}^i_j + \bar{\theta}_\textrm{bias}+\theta^i_\textrm{bias}$ where $\bm{Z}^i_j$ is the representation learned by previous layers. It is possible that $\bar{\theta}_\textrm{bias} + \theta^i_\textrm{bias}$ will correspond to a person's baseline level on the training set. As can be observed in \autoref{fig:bias}, $\theta^i_\textrm{bias}$ is highly correlated with the baseline level on all datasets, including IEMOCAP ($r=0.669$ for arousal and $r=0.543$ for valence). A potential explanation for why the magnitude of $\theta^i_\textrm{bias}$ is very small on IEMOCAP could be that the improvised dyads might be easier to predict, making it unnecessary for the model to encode the baseline levels.

\subsection{NME-CRF Experiments}

\textbf{NME-CRF improves performance:} We study the temporal structure of affective states on TPOT with the NME-CRF. While previous MLP-LME~\cite{xiong2019mixed} work does not generalize to temporal structures, such as modeled by a CRF, our NME easily extends CRFs. \autoref{tab:crf} shows that NME-CRF numerically improves over its baselines, demonstrating that even more complex models benefit from having person-specific parameters and that the transition patterns on TPOT depend on the person.

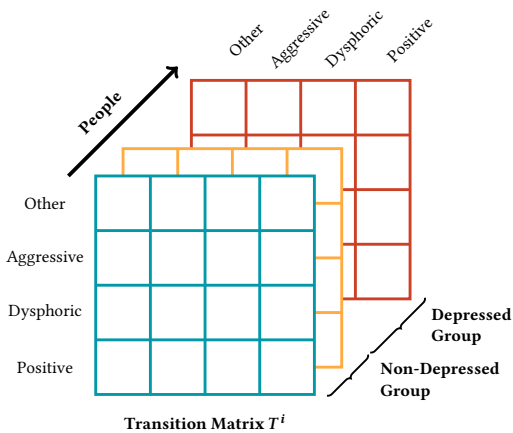
\begin{figure}
    \centering
    \resizebox{.8\linewidth}{!}{\begin{tikzpicture}
	\clip(-5.4,-4.4) rectangle (3.85,3.4);
	\begin{pgfonlayer}{nodelayer}
		\node [style=none] (0) at (-2, 1) {};
		\node [style=none] (1) at (-2, 0) {};
		\node [style=none] (2) at (-2, -2) {};
		\node [style=none] (3) at (-1, -2) {};
		\node [style=none] (7) at (2, 0) {};
		\node [style=none] (8) at (2, -2) {};
		\node [style=none] (9) at (1, -2) {};
		\node [style=none] (10) at (0, -2) {};
		\node [style=none] (13) at (-1, 2) {};
		\node [style=none] (14) at (-2, 2) {};
		\node [style=none] (15) at (0, 2) {};
		\node [style=none] (16) at (1, 2) {};
		\node [style=none] (17) at (2, 2) {};
		\node [style=none] (18) at (2, 1) {};
		\node [style=none] (61) at (-2, -1) {};
		\node [style=none] (62) at (2, -1) {};
		\node [style=none] (63) at (-3.25, -0.25) {};
		\node [style=none] (64) at (-3.25, -1.25) {};
		\node [style=none] (65) at (-3.25, -3.25) {};
		\node [style=none] (66) at (-2.25, -3.25) {};
		\node [style=none] (67) at (0.75, -1.25) {};
		\node [style=none] (68) at (0.75, -3.25) {};
		\node [style=none] (69) at (-0.25, -3.25) {};
		\node [style=none] (70) at (-1.25, -3.25) {};
		\node [style=none] (71) at (-2.25, 0.75) {};
		\node [style=none] (72) at (-3.25, 0.75) {};
		\node [style=none] (73) at (-1.25, 0.75) {};
		\node [style=none] (74) at (-0.25, 0.75) {};
		\node [style=none] (75) at (0.75, 0.75) {};
		\node [style=none] (76) at (0.75, -0.25) {};
		\node [style=none] (77) at (-3.25, -2.25) {};
		\node [style=none] (78) at (0.75, -2.25) {};
		\node [style=none] (79) at (-3.75, -0.75) {};
		\node [style=none] (80) at (-3.75, -1.75) {};
		\node [style=none] (81) at (-3.75, -3.75) {};
		\node [style=none] (82) at (-2.75, -3.75) {};
		\node [style=none] (83) at (0.25, -1.75) {};
		\node [style=none] (84) at (0.25, -3.75) {};
		\node [style=none] (85) at (-0.75, -3.75) {};
		\node [style=none] (86) at (-1.75, -3.75) {};
		\node [style=none] (87) at (-2.75, 0.25) {};
		\node [style=none] (88) at (-3.75, 0.25) {};
		\node [style=none] (89) at (-1.75, 0.25) {};
		\node [style=none] (90) at (-0.75, 0.25) {};
		\node [style=none] (91) at (0.25, 0.25) {};
		\node [style=none] (92) at (0.25, -0.75) {};
		\node [style=none] (93) at (-3.75, -2.75) {};
		\node [style=none] (94) at (0.25, -2.75) {};
		\node [style=none] (95) at (-1.75, -4.25) {\textbf{Transition Matrix} $\bm{T^i}$};
		\node [style=left] (96) at (3.125, -2.5) {\textbf{Depressed}\\\textbf{Group}};
		\node [style=left] (97) at (2.55, -3.5) {\textbf{Non-Depressed}\\\textbf{Group}};
		\node [style=none] (98) at (-4.25, 0.25) {};
		\node [style=none] (99) at (-2.25, 2.25) {};
		\node [style=rotate45] (100) at (-3.65, 1.4) {\textbf{People}};
		\node [style=rotate45] (101) at (-1, 2.75) {Other};
		\node [style=rotate45] (102) at (0, 2.75) {Aggressive};
		\node [style=rotate45] (103) at (1, 2.75) {Dysphoric};
		\node [style=rotate45] (104) at (2, 2.75) {Positive};
		\node [style=none] (105) at (-4.675, -0.25) {Other};
		\node [style=none] (106) at (-4.675, -1.25) {Aggressive};
		\node [style=none] (107) at (-4.675, -2.25) {Dysphoric};
		\node [style=none] (108) at (-4.675, -3.25) {Positive};
		\node [style=none] (109) at (0.25, -3.75) {};
		\node [style=none] (110) at (2, -2) {};
	\end{pgfonlayer}
	\begin{pgfonlayer}{edgelayer}
		\draw [style=solid3Fill] (14.center)
			 to (2.center)
			 to (8.center)
			 to (17.center)
			 to cycle;
		\draw [style=solid3] (13.center) to (3.center);
		\draw [style=solid3] (15.center) to (10.center);
		\draw [style=solid3] (16.center) to (9.center);
		\draw [style=solid3] (0.center) to (18.center);
		\draw [style=solid3] (7.center) to (1.center);
		\draw [style=solid3] (62.center) to (61.center);
		\draw [style=solid2Fill] (72.center)
			 to (65.center)
			 to (68.center)
			 to (75.center)
			 to cycle;
		\draw [style=solid2] (71.center) to (66.center);
		\draw [style=solid2] (73.center) to (70.center);
		\draw [style=solid2] (74.center) to (69.center);
		\draw [style=solid2] (63.center) to (76.center);
		\draw [style=solid2] (67.center) to (64.center);
		\draw [style=solid2] (78.center) to (77.center);
		\draw [style=solid1Fill] (91.center)
			 to (88.center)
			 to (81.center)
			 to (84.center)
			 to cycle;
		\draw [style=solid1] (87.center) to (82.center);
		\draw [style=solid1] (89.center) to (86.center);
		\draw [style=solid1] (90.center) to (85.center);
		\draw [style=solid1] (79.center) to (92.center);
		\draw [style=solid1] (83.center) to (80.center);
		\draw [style=solid1] (94.center) to (93.center);
		\draw [style=black arrow] (98.center) to (99.center);
	\end{pgfonlayer}
        \draw [decorate, decoration = {brace}, line width = 1pt] (1.23,-3.02) -- (0.48,-3.77);
        \draw [decorate, decoration = {brace}, line width = 1pt] (2.27,-1.98) -- (1.27,-2.98);
\end{tikzpicture}}
    \caption{Visualization of the person-specific transition matrices. Half of the matrices belong to families where the mother is in the depressed group.}
    \Description{Clarifying that transition matrices are from people which belong to groups.}
    \label{fig:transition}
\end{figure}

\textbf{Interpretation of temporal transitions:} The NME-CRF model allows analyzing the learned person-specific transition parameters. We focus on whether they differ between families (both adolescents and mothers) in the depressed and non-depressed group. We focus on this balanced group for two reasons 1) transition patterns have previously been linked to depression~\cite{schwartz2014parenting}, and 2) already the ground truth base rate of the four affective states is different between them as indicated by the Chi-squared test $\chi^2(3, 8946)=61.0, p<0.001$. As visualized in \autoref{fig:transition}, we group the person-specific transition matrices and then compare their differences. The multivariate Hilbert-Schmidt Independence Criterion (HSIC)~\cite{pfister2018kernel}\footnote{We use the implementation from the R package dHSIC.} indicates that the two groups have significantly different transition matrices $\textrm{HSIC}=0.71, p=0.006$.

The 95\% confidence intervals of the differences in the transition probabilities between families in the depressed and non-depressed group shown in \autoref{tab:transition} indicate six significant differences between them. While families in the non-depressed group are more likely to transition from \emph{positive} to the majority class \emph{other}, families in the depressed group are more likely to transition to \emph{aggressive} and \emph{dysphoric}. Similar trends are observed for transitions from \emph{other}: families in the non-depressed group are more likely to transition to \emph{positive} while families in the depressed group are more likely to transition into \emph{aggressive}. These observations seem plausible as more aggressive and less positive behaviors have been associated with depression~\cite{knox2000aggressive,schwartz2011observed,schwartz2014parenting}. As illustrated with the above analyses, it is possible to interpret the learned person-specific parameters learned by NME. 

\begin{table}
    \caption{95\% confidence intervals of the learned transition probability differences between families in the depressed and non-depressed group. Positive values indicate a higher transition probability for families in the depressed group. Intervals in bold are significantly different.}
    \center
    \resizebox{\linewidth}{!}{\begin{tabular}{
            c
            l
            >{{[}}
            S[table-format = -1.1,table-space-text-pre={[}]
            @{,\,}
            S[table-format = -1.1,table-space-text-post={]}]
            <{{]}}
            >{{[}}
            S[table-format = -1.1,table-space-text-pre={[}]
            @{,\,}
            S[table-format = -1.1,table-space-text-post={]}]
            <{{]}}
            >{{[}}
            S[table-format = -1.1,table-space-text-pre={[}]
            @{,\,}
            S[table-format = -1.1,table-space-text-post={]}]
            <{{]}}
            >{{[}}
            S[table-format = -1.1,table-space-text-pre={[}]
            @{,\,}
            S[table-format = -1.1,table-space-text-post={]}]
            <{{]}}
        }
        \toprule
        \multicolumn{2}{l}{Model-implied}& \multicolumn{8}{c}{Into}\\ \cmidrule(lr){3-10}
        \multicolumn{2}{l}{Transitions}& \multicolumn{2}{c}{Other}	& \multicolumn{2}{c}{Aggressive} & \multicolumn{2}{c}{Dysphoric} & \multicolumn{2}{c}{Positive} \\
        \midrule
        \multirow{4}{*}{\rotatebox[origin=c]{90}{From}} & Other	& 0.0 & 1.8 &  \bfseries{0.7} & \bfseries 4.9\mdseries & -2.0 & 3.2 & \bfseries{-7.4} & \bfseries -1.3\mdseries\\
        &Aggressive & -1.2 & 2.8 &-1.7& 0.2&-0.4& 3.4&\bfseries{-2.1}& \bfseries-0.4\mdseries\\
        &Dysphoric & -5.5 & 1.1&-0.1& 4.4&-0.9& 0.5&-1.4& 2.0\\
        &Positive & \bfseries{-8.3}& \bfseries-1.6\mdseries & \bfseries{0.3}& \bfseries 2.2 \mdseries& \bfseries{0.1}& \bfseries 5.5 \mdseries&0.0& 1.8\\
        \bottomrule
    \end{tabular}}
    \label{tab:transition}
\end{table}

\begin{figure*}
    \centering
    \resizebox{.9\linewidth}{!}{\input{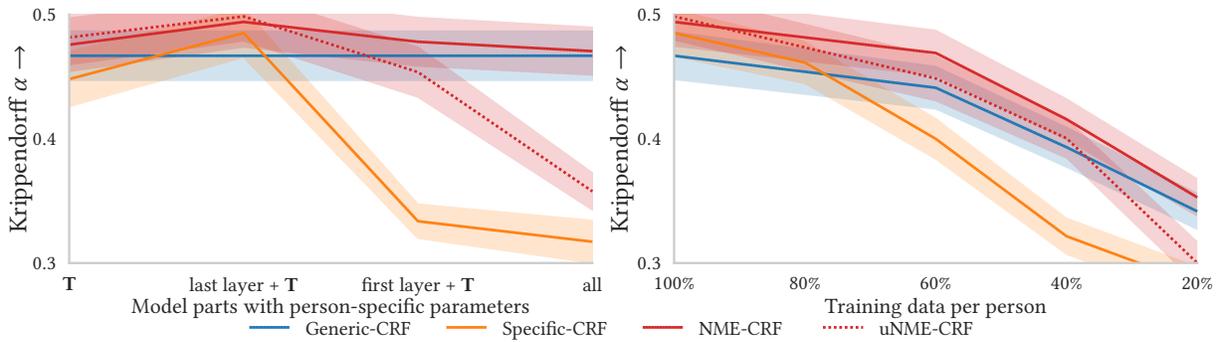}}
    \caption{Performance on TPOT: (left) with person-specific parameters in different model parts and (right) when trained on smaller subset of data per person.}
    \Description{Line plot of two ablation experiments.}
    \label{fig:unme}
\end{figure*}

\textbf{Regularization term needed for many person-specific parameters and small datasets:} To test in which situations the regularization term of NME, i.e., the right part of \autoref{eq:min}, is needed for good performance, we train an unregularized NME (uNME) that does not have the regularization term. We evaluate (u)NME with 1) person-specific parameters in different model parts of the CRF, and 2) with less and less training data per person. \autoref{fig:unme} indicates that the regularization term is needed for many person-specific parameters and on smaller datasets. Even with little data, NME-CRF always performs better than the Generic-CRF despite having more parameters. As described in \autoref{ssec:optimization}, mixed effect models tend to learn smaller person-specific parameters for a person with little data which helps avoid overfitting. In the extreme case of having very little data per person, the NME-CRF should converge to the Generic-CRF as the person-specific parameters will barely be used~\cite{pinheiro2000linear}. This trend can be observed in \autoref{fig:unme} as the performance gap between NME-CRF and Generic-CRF narrows with fewer observations per person.

\section{Conclusion}

We demonstrated that personalized models benefit by combining two types of trends: (a) person-generic trends shared across people and (b) unique person-specific trends. Linear mixed effect models are gaining popularity in machine learning for personalization as they combine these two trends. We proposed Neural Mixed Effect (NME) models to generalize previous work integrating linear mixed effect models in neural networks. NME allows person-specific parameters anywhere in a neural network to learn nonlinear person-specific trends. NME's optimization is further scalable to large datasets and large neural networks. NME achieved this by combining the efficient neural network optimization with the person-specific parameters of nonlinear mixed effect models. We evaluated NME on six unimodal and multimodal datasets covering regression and classification tasks and observed numerical improvements on all six datasets. Further, we showed that NME can be combined with neural conditional random fields to learn interpretable person-specific temporal transitions. Finally, we demonstrated that person-specific parameters can be interpreted, for example, we observed that the person-specific transition matrices of the NME-CRF are different for families in the depressed group. 

When multiple group variables are known to be present, e.g., people and different cultural backgrounds, it would be interesting to extend NME to a multilevel model~\cite{bosker2011multilevel}. An additional future direction, is evaluating which modalities, modal parts, or tasks benefit the most from NME.

\begin{acks}
This material is based upon work partially supported by Meta, National Science Foundation awards 1722822 and 1750439, and National Institutes of Health awards U01MH116923,  R01HD081362, R01MH125740, R01MH096951, R21MH130767 and R01MH132225. Any opinions, findings, conclusions, or recommendations expressed in this material are those of the author(s) and do not necessarily reflect the views of the sponsors, and no official endorsement should be inferred.
\end{acks}

\bibliographystyle{ACM-Reference-Format}
\bibliography{biblio}

\end{document}